\documentclass[10pt,journal,compsoc]{IEEEtran}
\usepackage{amsmath,graphicx}
\usepackage{amsfonts}
\usepackage{graphics}
\usepackage{latexsym}
\usepackage{float}
\usepackage{tabularx}
\usepackage[table,xcdraw]{xcolor}
\usepackage{multirow}
\usepackage{algpseudocode}
\usepackage{hyperref}
\usepackage{subfigure}
\usepackage{array}
\usepackage{enumitem}
\usepackage{color}
\usepackage{amsmath,bm}
\usepackage{subfigure}

\ifCLASSOPTIONcompsoc
  \usepackage[nocompress]{cite}
\else
  \usepackage{cite}
\fi

\ifCLASSINFOpdf
\else
\fi

\begin{document}
%
\title{IENet: Interactive Embranchment Network Based One-Stage Anchor Free Detector for Orientational Aerial Object Detection}
\author{Youtian~Lin,~
        Pengming~Feng,~\IEEEmembership{Member,~IEEE,}
        Jian~Guan,~\IEEEmembership{Member,~IEEE,}
        Wenwu~Wang,~\IEEEmembership{Senior~Member,~IEEE,}
        and~Jonathon~Chambers,~\IEEEmembership{Fellow,~IEEE}
\IEEEcompsocitemizethanks{\IEEEcompsocthanksitem Y. Lin and J. Guan are with the Group of Intelligent Signal Processing (GISP),  Harbin Engineering University, Harbin, China. E-mails: linyoutian.loyot@gmail.com, j.guan@hrbeu.edu.cn (Corresponding author: Jian Guan)
\IEEEcompsocthanksitem P. Feng is with the State Key Laboratory of Space-Ground Integrated Information Technology, CAST, Beijing, China. E-mail: p.feng.cn@outlook.com
\IEEEcompsocthanksitem W. Wang is with the Centre for Vision Speech and Signal Processing,  University of Surrey, Guildford, United Kingdom. E-mail: w.wang@surrey.ac.uk
\IEEEcompsocthanksitem J. A. Chambers is with the University of  Leicester, Leicestershire, United Kingdom. E-mail: jonathon.chambers@leicester.ac.uk}
%
}
\IEEEtitleabstractindextext{%
\begin{abstract}
Object detection in aerial images is a challenging task due to the lack of visible features and variant orientation of objects. Significant progress has been made recently for predicting targets from aerial images with horizontal bounding boxes (HBBs) and oriented bounding boxes (OBBs) using two-stage detectors with region based convolutional neural networks (R-CNN), involving object localization in one stage and object classification in the other. However, the computational complexity in two-stage detectors is often high, especially for orientational object detection, due to anchor matching and using   regions of interest (RoI) pooling for feature extraction. In this paper, we propose a one-stage anchor free detector for orientational object detection, namely, an interactive embranchment network (IENet), which is built upon a detector with  prediction in per-pixel fashion. First, a novel geometric transformation is employed to better represent the oriented object in angle prediction, then a branch interactive module with a self-attention mechanism is developed to fuse features from classification and box regression branches. Finally, we introduce an enhanced intersection over union (IoU) loss for OBB detection, which is computationally more efficient than regular polygon IoU. Experiments conducted  demonstrate the effectiveness and the superiority  of our proposed method, as compared with state-of-the-art detectors.
%
\end{abstract}

}
\maketitle
\textit{}

\IEEEdisplaynontitleabstractindextext

%
\IEEEpeerreviewmaketitle

\IEEEraisesectionheading{\section{Introduction}\label{sec:1}}

\IEEEPARstart{A}{utomatic} detection of visual objects is required in a variety of applications such as autonomous driving and robotics, where an object of interest in an image needs to be localized and recognized simultaneously. With the emergence of deep learning based techniques, significant progress has been made in object detection, especially for natural images.

Different from natural images, objects in aerial images are captured from bird's-eye view perspective, which often results in arbitrary object orientations, and leads to several significant challenges:
\begin{itemize}	
\item In aerial images, objects have more similar shape and fewer visible features than those in natural images (e.g., for houses and vehicles). This can lead to detection failure due to the challenges in distinguishing the similar shapes of different objects.
\item The highly complex background and variant appearances of targets increase the detection difficulties, especially for small and densely distributed targets.
\item The bird's-eye view perspective increases the complexity of the various orientations of objects, with concomitant  difficulty in obtaining the parameters to represent the angle diversity.	
\end{itemize}

Mainstream methods e.g.,  faster region-based convolutional  neural networks (Faster R-CNN)~\cite{ren2015faster}, you only look once (YOLO)~\cite{redmon2016you}, single shot multibox detector (SSD)~\cite{liu2016ssd}, originally developed for object detection in natural images, have been applied to address the above challenges. In these conventional detectors, horizontal bounding boxes (HBBs) are used for object detection. Although this representation performs well for natural images, it may lead to region overlap between objects in aerial images, especially  densely distributed targets with different rotations~\cite{xia2018dota, ding2018learning}. To address this problem, oriented bounding boxes (OBBs) are used as annotations which offer advantages in describing the bird's-eye view and rigid properties of objects~\cite{xia2018dota}.

Recently, several studies have been conducted for orientation detection, such as R-CNN based methods~\cite{liu2017rotated, ma2018arbitrary, zhang2018toward, ding2018learning, liu2016ship} and fully convolutional~\cite{xia2018dota, yang2018automatic, dai2016r} methods. These methods can be categorized approximately as two-stage methods and one-stage methods, respectively.

In two-stage detectors, the objects are detected in two stages, where the objects are first localized in the image, then they are classified, followed by location refinement. For example, in \cite{liu2016ship, xia2018dota}, the popular R-CNN \cite{ren2015faster} is adapted to perform orientation detection, using rotated regions of interest (RRoI) for orientation objects with rotated anchor boxes, leading to the so-called  rotation region-based convolutional neural networks (RR-CNN)  method. However, in the RR-CNN method, the object feature cannot be aligned when pooling is directly performed on the RRoI. In \cite{ding2018learning}, a learning based feature transformation method is used to align the orientation object features and to produce more precise object features. Although these RR-CNN based two-stage detectors show promising performance in detecting the orientation objects, the anchor matching and RoI pooling involved in these detectors are computationally expensive, which can result in low inference speed~\cite{lin2017focal}.

To improve the efficiency, fully convolution based one-stage detectors, such as \cite{tian2019fcos}, are adopted for orientation object detection using rotated anchor boxes. Such methods do not involve RRoI feature extraction, as a result, they are computationally more efficient than the two-stage methods, such as RR-CNN, but still offer competitive detection performance.

In the aforementioned two-stage and one-stage methods, anchor matching is required, where the intersection over union (IoU) between each anchor box and its ground truth needs to be computed, which incurs additional computational resources. Therefore, in this paper, inspired by  fully convolutional networks (FCNs)~\cite{long2015fully} and a fully convolutional one-stage (FCOS) detector~\cite{tian2019fcos}, we use an anchor-free one-stage architecture, e.g.~\cite{tian2019fcos, law2018cornernet, duan2019centernet, long2015fully} as our baseline method for orientation detection, where class probabilities and box regression coordinates are learned directly without the need for anchor matching, resulting in a new method, namely, a fully convolutional one-stage orientational (FCOS-O) object  detector.

In our proposed method, the original FCOS regression structure is extended by incorporating a separate branch for angle regression. In addition, a geometric transformation is proposed to represent the OBB by an HBB, with certain transformation parameters. As a result, this design enables FCOS for the detection of oriented objects. However, in this basic architecture, rotation angle cannot be associated with other properties such as the location of the target, thereby degrading the detection performance. To enhance the association, a novel method is developed, where an attention mechanism based procedure is employed for branch fusion, namely, an interactive embranchment (IE) module, where features from all the branches in the network are combined to enable the network to select consistent and relevant features for both rotated box regression and object classification. Moreover, in order to further enhance the training procedure, a simple OBB version of the IoU loss is introduced. We then show our proposed detector, IENet, outperforms the baseline one-stage detection method, and when compared with the state-of-the-art two-stage detectors, our model offers competitive performance, while improving computational efficiency.

\subsection{Summary of Contributions}
\label{sec:1_1}
Our novel contributions include:
\begin{enumerate}
	\item Proposing a one-stage anchor free detector for oriented object detection in aerial images, where a geometric transformation is introduced to substitute the OBB by an HBB and its corresponding orientation parameters. An OBB version of the IoU loss is given to regress the OBB for the target.
	\item Using an interactive embranchment (IE) module  to combine the orientational prediction task with both classification and box regression tasks, where a self-attention mechanism is employed to fuse the features from classification and regression branches, hence improving the accuracy of orientation detection.
	\item Showing  that our method outperforms the state-of-the-art detectors on public datasets for orientation detection in aerial images in terms of accuracy, computational complexity and memory efficiency.
\end{enumerate}

The remainder of the paper is structured as follows: In Section \ref{sec:Related_work}, both existing one and two stage orientational detectors are introduced, followed by brief discussion of the baseline method FCOS in Section \ref{sec:3_1}. Then the proposed IENet is described in Section \ref{sec:IENet}, including the novel geometric transformation, the self-attention mechanisms based interactive embranchment module and the OBB IoU loss. In Section \ref{sec:exp}, results and comparisons between the proposed approach and baseline methods are presented. Finally, Section \ref{sec:conclusion} provides a short conclusion and a discussion about the possible directions for future work.

\section{Related Work}
\label{sec:Related_work}
A variety of orientation detectors has been developed in the literature, which can be divided approximately into two categories based on  variation in their architectures, namely, two-stage and one-stage detectors. Moreover, aiming at orientation angle, innovations including self-attention mechanism have been developed to improve the accuracy of detectors.
\subsection{Two-Stage Detector}
\label{sec:2_1}
\begin{figure*}[t]
	\begin{center}
		\includegraphics[width=0.9\textwidth]{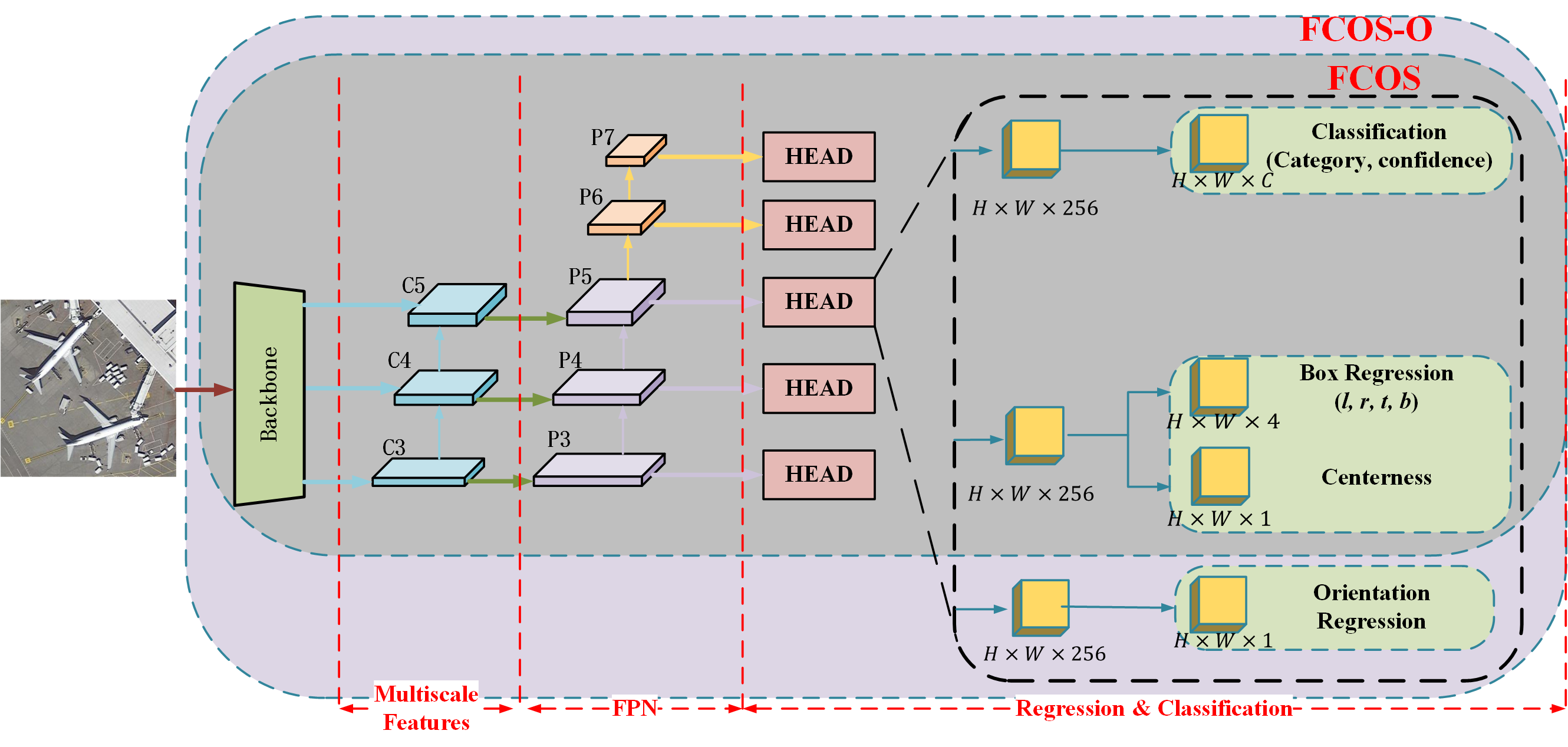}
	\end{center}
	\caption{The flowchart of the one-stage orientation baseline detector FCOS-O and the conventional FCOS, where $\mathbf{C}_{3}$, $\mathbf{C}_{4}$,  and $\mathbf{C}_{5}$, denote the feature maps from the backbone network and $\mathbf{P}_{3}$,  $\mathbf{P}_{4}$, $\mathbf{P}_{5}$, $\mathbf{P}_{6}$,   and $\mathbf{P}_{7}$  are the feature maps used for the final prediction. A backbone network is used to extract multiscale features for feature pyramid network (FPN)~\cite{lin2017feature}, then the shared prediction head is used for classification, box regression and the  orientation regression task. Different from conventional FCOS, an independent branch is employed to enable FCOS to regress the orientation parameters directly.}
	\label{model}
\end{figure*}

In two-stage detectors, object detection is achieved by inspecting the image twice, where the first inspection is to generate a region proposal set by detecting the possible regions that contain the object of interest (i.e., regions of interest, RoI), while the second inspection is to extract features using the backbone feature maps for each region proposal and passing these features to a classifier to identify the object category. One of the most popular two-stage methods on object detection is the R-CNN, introduced in \cite{girshick2015region}. Later, the Fast R-CNN~\cite{girshick2015fast} was proposed to improve the  R-CNN by designing an RoI pooling layer for the features to accelerate the processing speed.

Two-stage orientation detectors such as~\cite{zhang2018toward} address  orientation regression by adding anchors with different angles in both the region proposal and the RoI regression step, which allows the existing R-CNN based methods to produce an oriented bounding box by recognizing the object orientation angle. 
Recently, a rotated anchor is designed to generate an R-RoI, from which warping is used to extract a feature ~\cite{ma2018arbitrary, liu2017rotated}. However, with the R-RoI based method, many rotation proposals may be generated. According to \cite{zhang2018toward, azimi2018towards}, it is challenging to embed rotated proposal anchors in the network, due to the increased complexity for rotated proposal generation. 
In \cite{ding2018learning}, a method is proposed to avoid the rotated anchor computation by transforming the RoI to R-RoI using a light fully connected layer. They also add an IoU loss to match the two OBBs, which can effectively avoid the problem of misalignment. These two-stage detectors can obtain high detection performance at the cost of increased computational load.

Nevertheless, the feature extraction layers (e.g., RoI pooling~\cite{ren2015faster}, deformable convolution~\cite{dai2017deformable}, spatial transformer~\cite{jaderberg2015spatial}) in most R-CNN-based frameworks use horizontal bounding boxes  in the RoI pooling layer to extract corresponding object features, which is limited in predicting  oriented bounding boxes since the HBB contains more background information than the  OBB. As a result, this will lead to difficulties in extracting the overlapping features between objects. The RoI transformer based method \cite{ding2018learning} presents a solution for this problem by extracting the rotation region features for orientation objects.

These R-CNN based methods, however, rely on the RoI pooling that can be computationally expensive~\cite{lin2017focal}. In contrast, the one-stage detection methods can use an FCN to extract features and do not require RoI pooling to perform feature extraction, which can be computationally more efficient, as discussed next.
\subsection{One-Stage Detector}
\label{sec:2_2}

One-stage detectors~\cite{lin2017focal, tian2019fcos, liu2016ssd} directly predict the object category and location in a single-shot manner without any refinement step. Specifically,
one-stage detectors can be approximately classified into anchor-based methods, and anchor-free methods. In anchor-based methods, the HBB is predicted, and then transformed to an OBB, by simply adding the angle of the anchors. For example, in \cite{liao2018textboxes++}, a one-stage detector is presented for oriented scene text detection, where the HBB is used directly as an anchor to regress the OBB, and achieves state-of-the-art results on text detection. In anchor-based methods, the objects are detected by predicting the offsets with the dense anchor boxes, which could create a massive imbalance between positive and negative anchor boxes during training. Recently, a focal loss \cite{lin2017focal} was proposed to address this imbalance issue. Nevertheless, this still incurs substantial computational effort \cite{tian2019fcos}. In contrast, in anchor-free based detectors, the prediction is performed on a per-pixel manner, which frees the model from highly dense computation learning anchor matching.

Most one-stage oriented object detectors achieve high performance in the area of text scene detection~\cite{baek2019character, xing2019convolutional}, where masks are used to form the OBB \cite{xie2019scene}. These methods could be directly employed on aerial image datasets, where the objects are labeled with the OBB. However, text scene detection is very different from aerial object detection which has different challenges as mentioned in Section \ref{sec:1}.

On the other hand, many detectors down-sample the image to fit with the feature map size \cite{zhou2019objects, duan2019centernet, law2018cornernet}, and construct the final predicted object  by resizing the output object, which, however, can increase errors in the object detection, especially in the detection of objects from aerial images with densely distributed small targets. One idea to solve this problem is to predict an offset to reduce the resize error.

In the FCOS method \cite{tian2019fcos}, the object detection is achieved on a per-pixel fashion. In this way, the detection errors caused by resizing the image are avoided since the key points on the output feature maps  correspond to a pixel coordinate in the input image. Inspired by this idea,  the structure of FCOS is employed as the baseline architecture in our proposed method.

 However, object detection in aerial images is different from that of scene text detection, and it is a more challenging task due to the sizeable dense cluster object having more misalignment features. To the best of our knowledge, until now, there is no anchor free solution in one-stage oriented object detection for aerial images, since  one-stage detectors cannot extract features in the OBB as in the RoI pooling step in two-stage detectors. As such, it is essential to build an appropriate feature extractor for the detector to recognize the object orientation in one stage anchor-free orientational detectors.

Therefore, in our proposed IENet, an anchor free one-stage framework is employed to directly predict the object without the complex computation induced by anchor matching and RoI feature extraction.

\subsection{Self-Attention Mechanism}
\label{sec:2_3}
The self-attention mechanism~\cite{vaswani2017attention, bahdanau2014neural} was originally proposed to solve the machine translation problem which is used to capture global dependencies. Recently, self-attention has been applied in computer vision tasks ~\cite{buades2005non, parmar2018image} to capture interrelated features for CNN based detectors. Furthermore, in \cite{lafferty2001conditional, xu2015show, yang2016stacked}, non-local operation is used for capturing long-range dependencies which achieves state-of-the-art classification accuracy. Moreover, the self-attention mechanism is applied for object detection and instance segmentation, and achieves high mean average precision (mAP) \cite{huang2019interlaced}.

In this work, a self-attention mechanism is designed within the IE module to find the relationship between the feature maps from each branch, thereby identifying the most suitable features for OBB regression.
\section{Baseline Method}
\label{sec:3_1}
Based on the FCOS method, we present a baseline architecture by adding an orientation regression branch, namely FCOS-O, as illustrated in Fig.~\ref{model}. This architecture allows the model to solve orientation detection in a per-pixel manner, which can avoid  resizing errors. Our proposed IENet is designed based on this architecture to improve the orientation detection performance, which will be described in the next section.

As shown in Fig.~\ref{model}, in the FCOS architecture, a CNN backbone is employed to extract features from the input images, followed by an FPN \cite{lin2017feature}, where different levels of feature maps are obtained. Let $\mathbf{C}_{i}$ be the feature maps of the backbone network and $\mathbf{P}_{i}$ be the feature levels obtained by the FPN, where $i$ indicates the layer index of the feature map. In this work, following FCOS \cite{tian2019fcos}, five levels of feature maps ${\mathbf{P}_{3}, \mathbf{P}_{4}, \mathbf{P}_{5}, \mathbf{P}_{6}, \mathbf{P}_{7}}$ are used, where $\mathbf{P}_{3}$, $\mathbf{P}_{4}$ and $\mathbf{P}_{5}$ are produced from the backbone CNN's feature maps $\mathbf{C}_{3}$, $\mathbf{C}_{4}$ and $\mathbf{C}_{5}$, respectively, and are top-down connected by a 1$\times$1 convolutional layer, while $\mathbf{P}_{6}$ and $\mathbf{P}_{7}$ are produced by applying convolutional layers with stride of size 2 on $\mathbf{P}_{5}$ and $\mathbf{P}_{6}$, respectively. 
After obtaining the feature maps from the FPN, a shared prediction ``$\text{HEAD}$" is applied to predict the object category and spatial location for each feature map. Here, each feature map contains feature information from different levels of convolutional processes.

In the shared prediction ``$\text{HEAD}$", two separate branches are employed for detection, namely, classification and box regression, 
respectively. Following \cite{tian2019fcos}, four convolutional layers are added for each branch, and $C$ binary classifiers are trained for classification.  Let $\mathbf{F}_{i} \in \mathbb{R}^{W\times H\times C}$ be the feature maps with size $(W, H)$ at layer $i$ of the network, $s$ be the total stride until the $i$-th layer, and $C$ be the number of categories. Each keypoint on feature map $\textbf{F}_{i}$ with location $\mathbf{p}_s=(x_s, y_s)$ can be mapped back onto the input image via
\begin{equation}
\begin{array}{ll}{x = \lfloor \frac{s}{2} \rfloor + x_s } \\ {y = \lfloor \frac{s}{2} \rfloor + y_s}\end{array}
\label{image-location}
\end{equation}
where $(x, y)$ is the location on the input image and $\lfloor \cdot \rfloor$ denotes the round-down operator. Here, $\mathbf{p}_s$ is considered as a positive sample if it falls into any ground truth box within a radius $d$ of the box center and belongs to the class with label $c$. Otherwise, it is a negative sample with $c=0$, which denotes the background.

\begin{figure}[hb!]
	\begin{center}
	\subfigure[]{
	\includegraphics[width=0.23\textwidth]{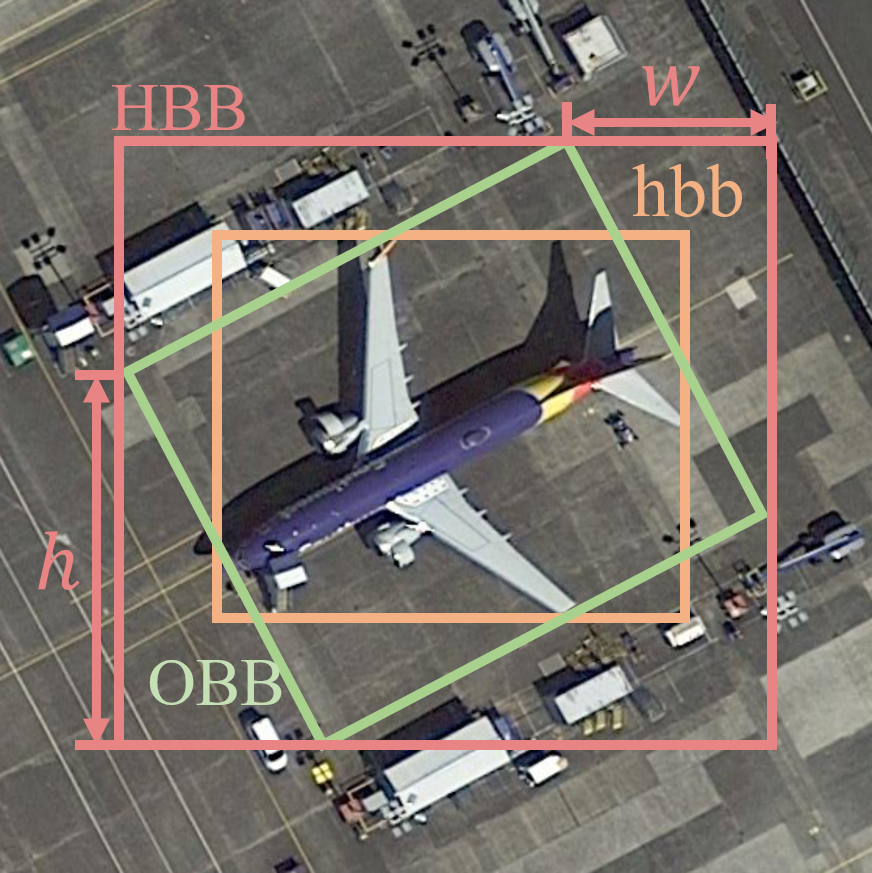}}
	\subfigure[]{
	\includegraphics[width=0.23\textwidth]{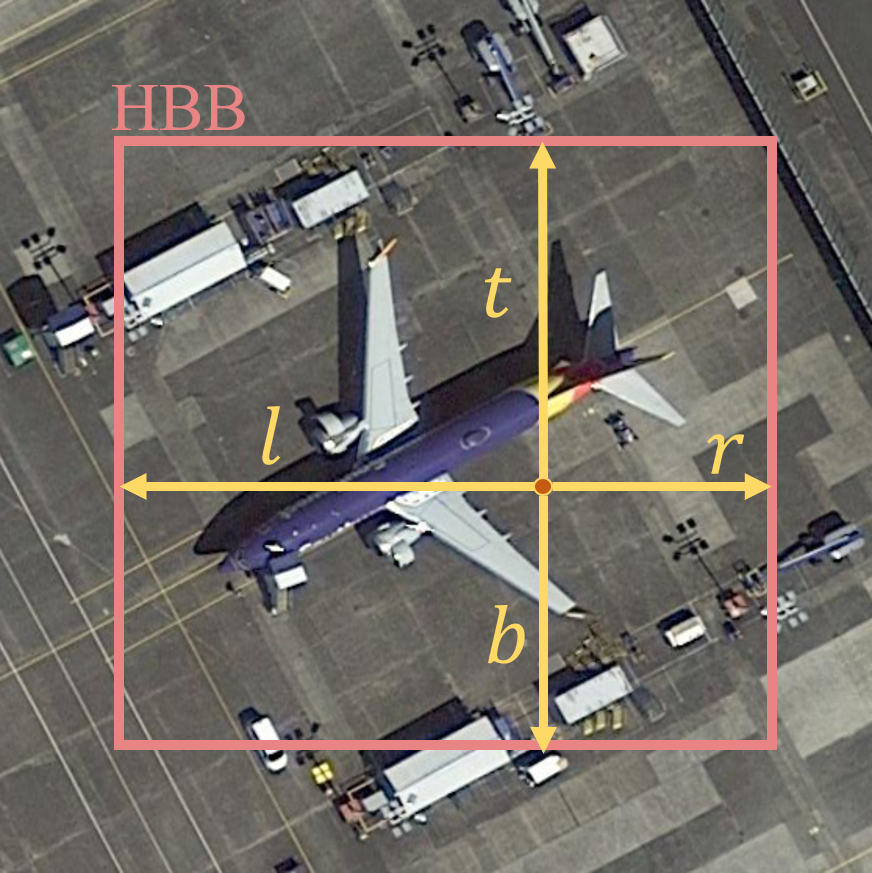}}
	\end{center}
	\caption{Geometric transformation and regression parameters used in this work. (a) is the geometric transformation method employed for  processing the oriented bounding box (OBB) of the target to its surrounding horizontal bounding box (HBB), where $h$ and $w$ are the transformation parameters. Different from HBB used in this work, hbb is the inner compact horizontal bounding box usually given as horizontal bounding box ground truth. (b) shows the 4D regression vector $[l, t, r, b]$ employed  to encode the location of an HBB for each foreground pixel.}
	\label{HBB_to_OBB}
\end{figure}

Meanwhile, in the box regression branch, as shown in Fig. \ref{HBB_to_OBB} (b), for each location on  the feature map, FCOS employs a 4D vector  $[l, t, r, b]$ to regress an HBB. Given a keypoint $(x_s, y_s)$ and its surrounding HBB with left-top and bottom-right pixels  $[x_{smin}, y_{smin}, x_{smax}, y_{smax}]$ on the feature map, keypoint based regression can be formulated as
\begin{equation}\label{lrtb}
  \begin{aligned}
    l = x_{s}-x_{smin}, &  &t = y_{s}-y_{smin} \\
    r = x_{smax}-x_{s}, &  &b = y_{smax}-y_{s}
   \end{aligned}
\end{equation}

Moreover, in the box regression branch, a simple but effective strategy is used to fine-tune the center of the bounding box $(x,y)$, which depicts the normalized distance from the location to the center of the object.  Given the regression targets $l,r, t$ and $b$ for a location $(x, y)$, the centerness can be computed as
\begin{equation}
\label{eq:10}
  \mbox{centerness} = \sqrt{\frac{min(l,r)}{max(l,r)}\times \frac{min(t,b)}{max(t,b)}}
\end{equation}
where $\sqrt{\cdot}$ is the square root, which is used to slow down the decay.
A standard binary cross entropy (BCE) loss is employed to calculate the centerness, whose value ranges from 0 to 1.

The above represents the pipeline of the FCOS detector, however, this architecture cannot directly predict the OBB for the object. In order to address this limitation, we construct a baseline architecture FCOS-O based on FCOS by adding a new regression branch (i.e., orientation regression) in the shared prediction ``$\text{HEAD}$", as shown in  Fig~\ref{model}. In this new architecture, the OBB of the oriented object can be represented as an HBB from the box regression branch with an angle prediction from the orientation regression branch.

Note that, our proposed IENet is designed based on the FCOS-O detector, where the main difference between the proposed IENet and FCOS-O is the design of the prediction head. In the next section, the proposed IENet will be described, where the representation of the oriented bounding box, IE module and the loss function will be given in detail.

\section{Proposed IENet}
\label{sec:IENet}
\begin{figure*}[htb!]
	\begin{center}
		\includegraphics[width=0.9\linewidth]{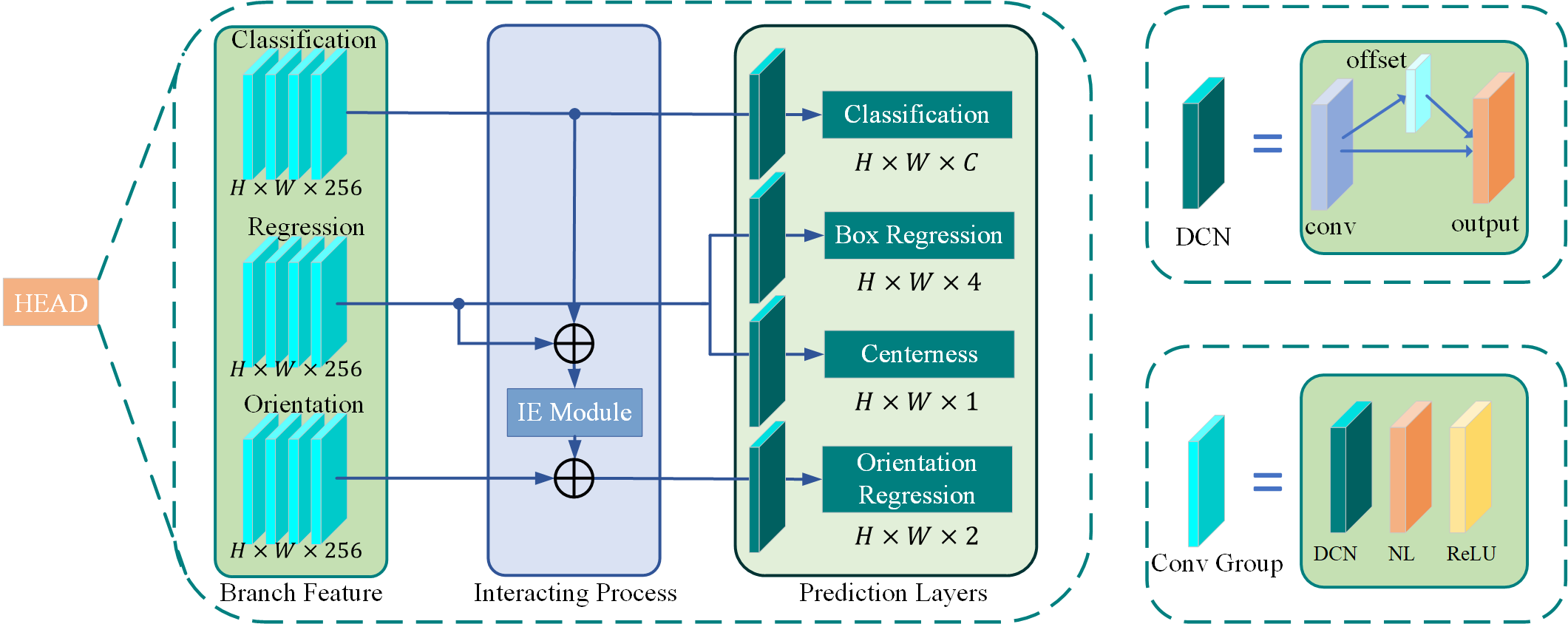}
	\end{center}
	\caption{Feature head for prediction in IENet, which contains three independent branches, namely the classification, regression and orientation branches. A downstream resampling procedure constituted by the four convolution group is firstly used on each branch, and each convolutional group contains a convolutional layer, normalization layer and a ReLU layer. A self attention based IE Module is firstly employed to merge features from the classification and the regression branch, then  orientation feature maps are composed by adding features  from the original orientation branch. $W, H$ indicate the downstream output sizes from the backbone network, $C$ is the number of categories, and $\bigoplus$ denotes the element-wise addition.}
	\label{model_head}
\end{figure*}

In this section, the proposed detector, IENet, is described in detail. Firstly, a novel representation for the  OBB is presented in Section \ref{sec:4_1}. Then, the network architecture with self-attention mechanism based IE module is presented in Section \ref{sec:4_2}. In Section \ref{loss}, the loss function used in the model is discussed. Finally, details for parameter inference are provided in Section \ref{inference}.  The prediction head of our proposed IENet is illustrated in Fig.~\ref{model_head}.
\subsection{Representation of the Oriented Bounding Box}
\label{sec:4_1}
Different from the rotation anchor based orientation detectors, which aim to predict the OBB of the target, we use the HBB surrounding a target and the transformation parameters to describe the OBB of the target. As a result, a more effective one-stage key-point based method \cite{duan2019centernet} can be employed directly for target presentation and prediction. As shown in Fig.~\ref{HBB_to_OBB} (a), in the training process, the ground truth OBB is firstly transformed to its surrounding HBB with transformation parameters $h$ and $w$, which is easier to be regressed.  In more detail, as shown in Fig.~\ref{HBB_to_OBB} (a), starting from the top-left point, an 8-dimensional vector $[x_{1}, y_{1}, x_{2}, y_{2}, x_{3}, y_{3}, x_{4}, y_{4}]$ is used to include the four pixels of a ground truth OBB, its surrounding HBB with top-left and bottom-right pixels $[x_{min}, y_{min}, x_{max}, y_{max}]$ is used for prediction, together with the orientation parameters $[w, h]$, which are calculated as:
\begin{equation}\label{wh}
  \begin{aligned}
    w = x_{max}-x_{2} \\
    h = y_{max}-y_{1}
   \end{aligned}
\end{equation}
In this way, the problem of orientation detection can be simply addressed by predicting the HBB with its corresponding transformation parameters, and can be solved with a one-stage detection framework.

After representing the OBB with its surrounding HBB and orientation parameters, the task of OBB regression can be divided into HBB regression and orientation regression separately, where the HBB is constructed following~\cite{tian2019fcos} as shown in Fig.~\ref{HBB_to_OBB} (b). The box regression vector $R_b = [l, t, r, b]$ denotes the offset that is calculated between the regression point and its four HBB boundaries,  where $l, t, r, b$ represent the left, top, right, and bottom distances, respectively. Meanwhile, the orientational angle is converted to an orientation regression vector $R_o = [w, h]$ following Eq. (\ref{wh}) as shown in Fig.~\ref{HBB_to_OBB} (a). In this way, the original OBB is represented by a 6-dimensional vector, $[l, t, r, b, w, h]$, which makes it easier for regression. Note that, in this paper,  the HBB is the extended box of the OBB, which is used for box regression.

In the next section, the network architecture is designed for orientation object detection based on the above geometric transformation.

\subsection{Interactive Embranchment Module}
\label{sec:4_2}
In order to address the problem of orientation regression, most CNN based  detectors \cite{liu2019omnidirectional, liu2019tightness, liao2018textboxes++} employ an independent branch directly from feature heads as the orientation branch, i.e., as shown in Fig. \ref{model_head}, which is separate from the classification and the box regression branch. In our work, we present a geometric transformation to calculate the orientation parameters $[w, h]$ associated with the HBB, therefore, the relationship between the regression and the  orientation branch can be employed to enhance the orientation regression.

Moreover, as shown in Fig. \ref{model_head}, a connection is built to establish the cooperation between classification, box regression and orientation regression branches. Firstly, in order to achieve more reliable features, following \cite{duan2019centernet}, we use a deformable convolution network (DCN)~\cite{dai2017deformable} on each branch, and then add the feature maps from classification and regression branches together. Furthermore, an interactive embranchment module, namely the IE module, is employed for feature fusion. Then the output feature maps from the IE module are added to the feature maps from the orientation regression branch and yield the final feature maps for orientation regression, which will be detailed in the next section. In this way, relationship between the three independent branches is built interactively, while the original information describing the feature maps is retained. With this idea, we can improve the performance of orientation regression. In addition, to allow the prediction head  to capture more consistent features in the entire process of the IE module, all the convolutional layers used in the prediction head are constructed with $3 \times 3$ kernels.

\begin{figure}
	\begin{center}
		\includegraphics[width=.98\linewidth]{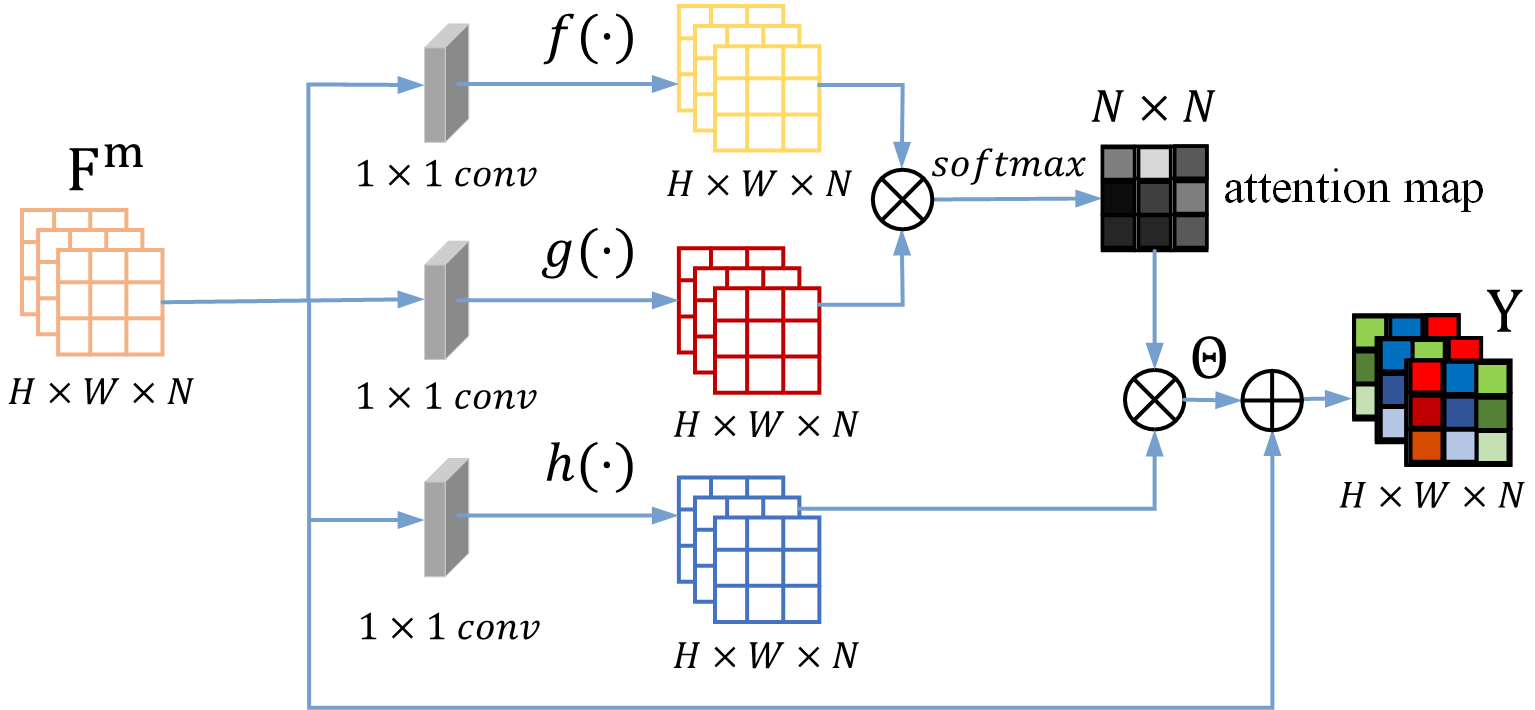}
	\end{center}
	\caption{Self-attention mechanism based IE module, where $\bigotimes$ indicates the matrix multiplication, and  $\bigoplus$ denotes the element-wise addition.}
	\label{self-attention}
\end{figure}
\label{IEBlack}
To further improve the oriented prediction accuracy, a self-attention module is employed to fuse the features from both the classification branch and the regression branch. This is different from \cite{zhang2018self, wang2018non}, where the self-attention module is applied for a single branch feature. As the self-attention can establish the relationship between those feature maps, more flexible features can be selected for orientation regression.

As shown in Fig. \ref{self-attention}, after obtaining the merged feature maps $\textbf{F}^{m}$ with $N$ channels from classification and box regression branches, the self-attention module is constructed by using three $1 \times 1$ convolutional layers and a softmax layer. Firstly, the features are projected to three feature spaces via three different convolutional layers $f(\textbf{F}^{m}), g(\textbf{F}^{m}), h(\textbf{F}^{m})$, where $f(\cdot)$ together with $g(\cdot)$ form an attention map via the softmax function. Then the attention map can be used to indicate the relationship among the input features and yield a retroaction on $h(\cdot)$.
Here, $h(\cdot)$ is used to obtain  the original input feature maps. Specially, the attention map $\mathbf{\Upsilon}$ can be calculated as
\begin{equation}
\label{eq:4}
\mathbf{\Upsilon} = softmax(f\left(\textbf{F}^{m}\right)^{T}g\left(\textbf{F}^{m}\right))
\end{equation}
where $f\left(\textbf{F}^{m}\right)^{T}g\left(\textbf{F}^{m}\right)$ outputs an $N \times N$ matrix $\Xi$. Here,  the $softmax$ function is applied to each element of $\Xi$ to build the relationship between feature maps, which is expressed as follows
\begin{equation}
\label{eq:5}
\mathbf{\Upsilon}_{q, p}=\frac{\exp \left(\Xi_{pq}\right)}{\sum_{p=1}^{N} \exp \left(\Xi_{pq}\right)}
\end{equation}
where $p$ and $q$ are the row and column indices of matrix $\Xi$, respectively. Here, $q$ and $p \in \{1, \cdots ,N\}$.  In this way, we can obtain the attention layer  $\Theta=\left(\theta_{1}, \theta_{2}, \cdots, \theta_{q}, \cdots, \theta_{N}\right)$, where
\begin{equation}
\label{eq:6}
\theta_{q}=\sum_{p=1}^{N} \mathbf{\Upsilon}_{q, p} h\left(\textbf{F}^{m}_{p}\right)
\end{equation}
In addition, to maintain the original features while avoiding the gradient vanishing problem, a short connection is built between the input feature maps and $\Theta$, following~\cite{wang2018non}, the final output of the attention layer  is performed by a convex weight $\gamma$, denoted as
\begin{equation}
\label{eq:7}
\mathbf{Y}=\gamma \Theta+\textbf{F}^{m}
\end{equation}

In this way, the self-attention mechanism based IE module is built, with which the feature maps from different branches are fused to generate more reliable features. In this work, $\gamma$ is simply set to 1 which can relate the original feature information to the output. Moreover, $\mathbf{Y}$ is the IE module output feature maps, which will be directly added to the orientation branch, and yields the final orientation feature maps for orientation prediction.

\subsection{Loss Function}
\label{loss}
To train the network, we use a loss function calculated over all locations on the feature maps as follows:
\begin{equation}
\label{eq:8}
L=\frac{1}{N_{p o s}} L_{c l s}+\frac{\lambda}{N_{p o s}} L_{reg}+\frac{\omega}{N_{p o s}} L_{ori}
\end{equation}
where $N_{pos}$ is the number of positive targets in the ground truth, and $\lambda$, $\omega$ are the scale coefficients used to reduce $L_{reg}$ and $L_{ori}$, respectively, to balance the loss. $L_{cls}$, $L_{reg}$ and $L_{ori}$ denote the loss from classification, box regression and orientation regression, respectively, wherein $L_{cls}$ is calculated on each pixel by the focal loss function~\cite{lin2017focal},
\begin{equation}
\label{eq:9}
  L_{cls} = \frac{-1}{M}\sum \left\{
  \begin{aligned}
	\alpha(1-\hat{Y}_{cls})^{\beta}log(\hat{Y}_{cls})&  & {Y_{cls}=1} \\
	\alpha(\hat{Y}_{cls})^{\beta}\times log(1-\hat{Y}_{cls})&  & \mbox{otherwise}
  \end{aligned}
  \right.
\end{equation}
where $Y_{cls}\in[0,1]$ is the ground truth for classification and $\hat{Y}_{cls}$ is the classification score from the network, $\alpha$ and $\beta$ are hyper-parameters of the focal loss. $M$ is the number of selected keypoints in the image, which is used to normalize all positive focal loss instances to 1.
In order to reduce the computational complexity caused by computing $\text{IoU}_{\text{OBB}}$, i.e. the orientation IoU, an inner box is introduced in our IENet to calculate $\text{IoU}_{\text{OBB}}$ instead of OBB, which can be obtained  by encoding $[l,r, t, b]$ and their corresponding orientation parameters $[w, h]$, as follows
\begin{equation}
\label{eq:11}
\begin{array}{ll}{l_o= \lvert l-w \rvert, } & {t_o=\lvert t-h \rvert, } \\ {r_o=\lvert r-w \rvert,} & {b_o=\lvert b-h \rvert.}
\end{array}
\end{equation}
where $[l_o, t_o, r_o, b_o]$ denotes the inner box offset, which is used to calculate the IoU loss for OBB.

In this way, the box regression loss $ L_{reg} $ can be calculated as
\begin{equation}
\label{eq:12}
\begin{aligned}
L_{reg}=&\text{BCE}(\widehat{centerness}, centerness) \\
&+\lambda_{reg} Smooth_{L1}(\hat{R_b}, R_b)  \\
&+(1 - \text{IoU}_{\text{HBB}})
\end{aligned}
\textit{}\end{equation}
where $\lambda_{reg}$ is a penalty parameter to prevent the disparity magnitude between the losses. Here, $\hat{\cdot}$ denotes the  prediction result. The loss $L_{ori}$ from the  orientation branch  is formulated as follows
\begin{equation}
\label{eq:13}
\begin{aligned}
L_{ori}&= \lambda_{ori} Smooth_{L1}(\hat{R_o}, R_o) \\
&+ (1 - \text{IoU}_{\text{OBB}})
\end{aligned}
\end{equation}
Here, $\lambda_{ori}$ is the penalty parameter, and $\text{IoU}_{\text{HBB}}$ is the intersection over union for the HBB following \cite{tian2019fcos}, and $\text{IoU}_{\text{OBB}}$ is computed in the same way. Thus the parameters $[l, t, r, b]$ and $[w, h]$ are combined to transform the HBB to the OBB.

So far, the IENet framework has been constructed and the prediction can be achieved following the given processes as above.
\subsection{Inference}
\label{inference}
The inference of the IENet parameters is straightforward. Given an input image, we use the output from the last three layers of the backbone network as the input of the FPN, which fuses three feature maps and generates the final feature maps for the prediction head, as illustrated in Fig.~\ref{model}. Each prediction head contains three branches, and each branch is designed for a different task, i.e., object classification, box regression for bounding box prediction, and orientation parameters estimation, respectively, as illustrated in Fig.~\ref{model_head}. Here, the prediction head is shared among the five feature maps. The size of the head map is the same as the feature maps generated by the backbone network, therefore the location on each prediction map $\textbf{F}$ can be projected to a location on the image according to  Equation \eqref{image-location}, for each location, we select those whose classification confidences are higher than 0.5 as the definite prediction. However, we use predicted centerness to multiply the predicted category, therefore the threshold is set as 0.05. Finally, the proposed IENet can predict the 4-dimensional vector $[l, t, r, b]$ and the 2-dimensional vector $[w, h]$, and then these parameters can be transformed to the OBB as discussed in Section \ref{sec:4_1}.
\section{Experiments and Results}
\label{sec:exp}
In this section, experiments are provided to examine the performance as well as the efficiency of our proposed IENet, and to compare with results from other state-of-the-art detectors.
\subsection{Datasets and Evaluation Measurements}
\label{sec:datasets}
We evaluate our proposed IENet on two challenging datasets, known as DOTA and HRSC2016~\cite{lb2017high}. Both datasets contain aerial images with a mass of objects with arbitrary orientations. The ablation studies are conducted on DOTA, and all results are provided by the evaluation servers.

\begin{itemize}
\item \textbf{DOTA} is the largest dataset for object detection in aerial images with oriented bounding box annotations. 2806 aerial images are collected from different sensors and platforms. Each image is of  size around $4000\times4000$ pixels and contains objects exhibiting a wide variety of scales, orientations, and shapes. These DOTA images are then annotated by experts in aerial image interpretation using 15 common object categories. The fully annotated DOTA images contain 188282 instances, including Baseball diamond (BD), Ground track field (GTF), Small vehicle (SV), Large vehicle (LV), Tennis court (TC), Basketball court (BC), Storage tank (ST), Soccer-ball field (SBF), Roundabout (RA), Swimming pool (SP), and Helicopter (HC).

\item \textbf{HRSC2016} is a challenging dataset for ship detection in aerial images. The images are collected from Google Earth. It contains 1061 images and more than 20 categories of ships in various appearances. The image size ranges from $300 \times 300$ to $ 1500 \times 900$. The training, validation and test set include 436 images, 181 images and 444 images, respectively.
\end{itemize}

In this work, the performance metrics, i.e., the average precision (AP) of all categories and mAP over all categories, are employed for evaluation and comparison with state-of-the-art methods. Here, the AP scores are calculated following the Pascal Visual Object Classes (VOC) challenge~\cite{everingham2015pascal}, which is drawn based on Precision and Recall. Moreover, Precision and Recall are calculated using IoU between the predicted box and ground truth box with classification accuracy as a performance metric.

\begin{figure*}
	
	\begin{center}
		\includegraphics[width=0.85\textwidth]{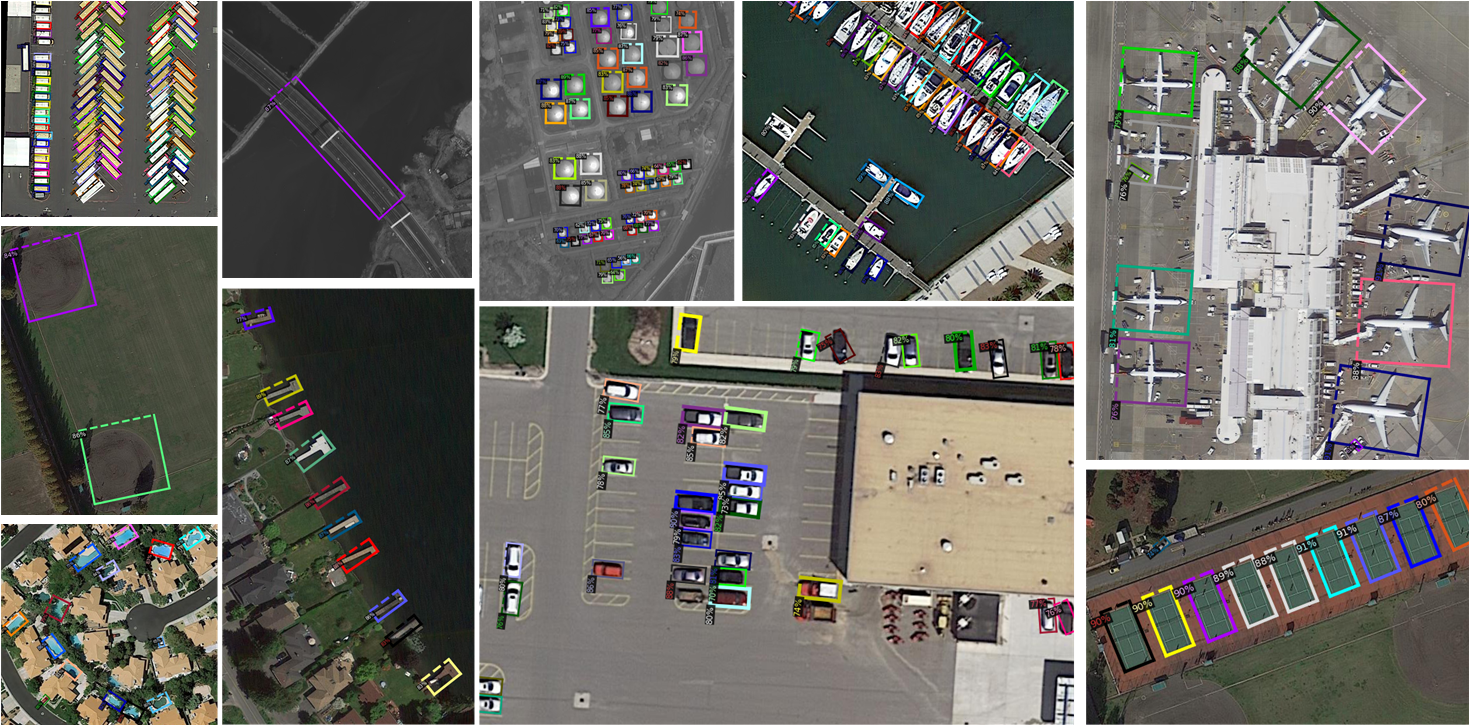}
	\end{center}
	\caption{Illustration of the proposed IENet on the DOTA dataset in different scenarios, selected categories `large vehicle', `bridge', `storage tank', `ship', `plane', `baseball field', `swimming pool', `harbor', `small vehicle' and `tennis court' are shown in sequence and the oriented objects are correctly detected.}\label{DOTA_results}\label{figure:5}
\end{figure*}

\begin{figure*}
	
	\begin{center}
		\includegraphics[width=0.85\textwidth]{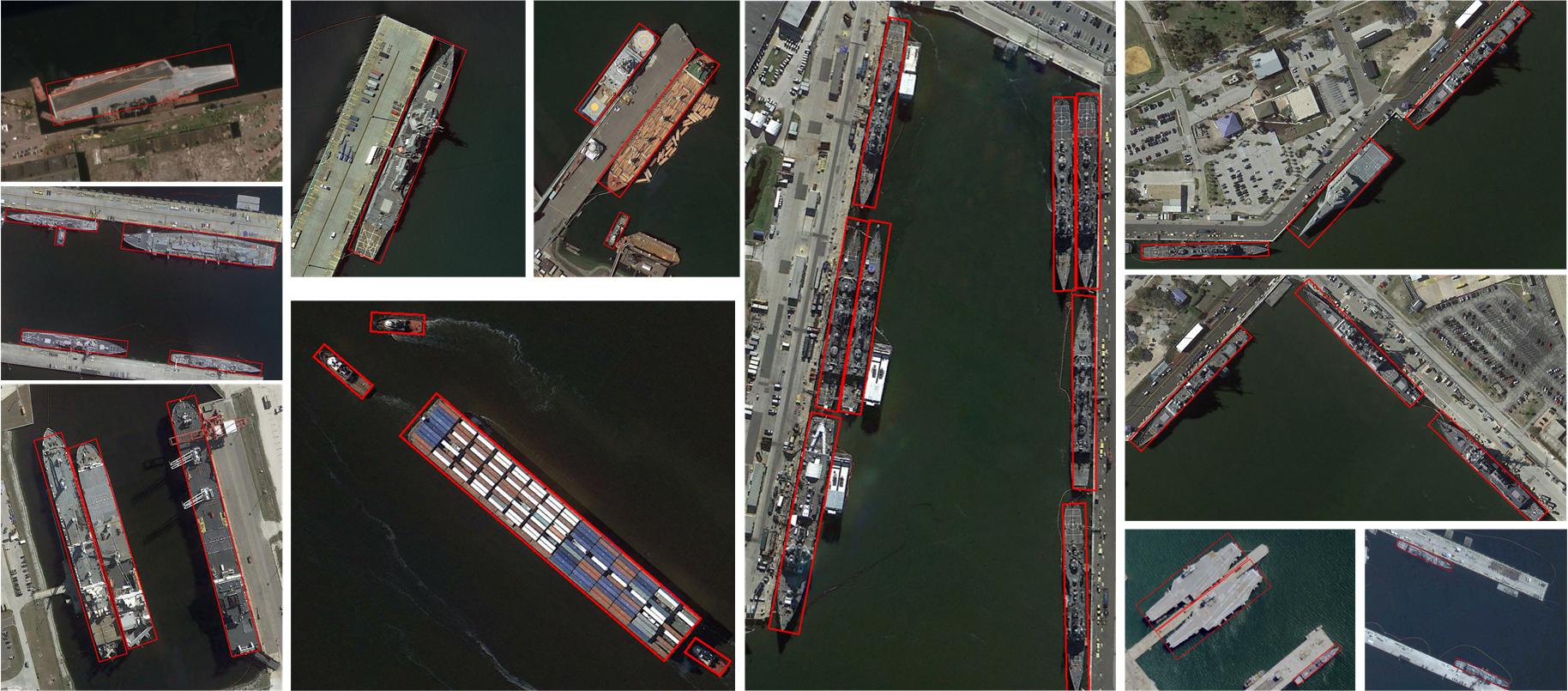}
	\end{center}
	\caption{Illustration  of the proposed IENet on the HRSC2016 dataset, where different kinds of ships are correctly detected with their orientation angles.}\label{hrsc_results}\label{figure:6}
\end{figure*}

\subsection{Implementation Details}

\textbf{Datasets Settings} In our experiment, DOTA is used by following \cite{ding2018learning}, where both training and validation sets are used for training, and the test set is employed for testing \cite{xia2018dota}. Only limited data augmentation is employed, specially,  we resize the images at the scale of 0.5, and use both original images and  resized images for training and testing. After image resizing, following~\cite{ding2018learning}, all images from datasets are cropped to \(1024 \times 1024\) pixels with a stride of 824 for memory efficiency. For those categories with few samples, a rotation augmentation is randomly adopted from 4 angles (i.e., 0, 90, 180, 270 degrees) to avoid the effect of imbalance between different categories. For the evaluations on the HRSC2016 dataset, we only apply horizontal flipping. The images are resized to \(512 \times 800\) pixel images, where 512 represents the length of the short side and 800 is the maximum length of an image. Moreover, we refine the  targets with 90 degrees and 180 degrees to
 -90 degrees and  -180 degrees respectively, which helps avoid the confusion caused by object orientation,  since the targets with 90 and -90 degrees or the targets with 180 and -180 degrees share the same posture.

\textbf{Network Settings} To compare with other state-of-the-art methods,  we use ResNet-101~\cite{he2016deep} as the backbone network in all the experiments. Then, the convolutional weight that composes the prediction head is initialized with normal distribution, where the standard deviation is set to be  0.01. Also, the convolutional bias is enabled and initialized with 0. We use the group normalization~\cite{wu2018group} in the prediction head, which can significantly outperform the batch normalization~\cite{ioffe2015batch} for small batch size. We set $\lambda$ and $\omega$ in the loss function as 1, and the L1 loss scale coefficient $\lambda_{reg}$ and $\lambda_{ori}$ as 0.2. Furthermore, we tune the focal loss hyper-parameters to achieve  better adaptability with DOTA and HRSC2016 datasets, for which $\alpha$  and $\beta$ are set to be 0.3 and 4.0, respectively. Note that, this network setting is empirically achieved, which has been found to yield good performance in our experiments.

\textbf{Training Details} Our network is trained using 8 GTX 1080 Ti GPUs, and each has 12GB memory. Here, the batch size is set as 8 due to the GPU memory limitation. In addition, to stabilize the training process and yield better performance, we adopt a learning rate schedule with a combination of a warm-up training and a multi-step learning rate. Here, a linear warm-up method is applied, where the warm-up factor is set as 0.001 with 1000 iterations in the earliest training stage, and the learning rate is initialized with 0.01 but reduced by a factor of 10 at the end of the last 20K training step (i.e., from 80K to 100K). Furthermore, we use stochastic gradient descent (SGD)~\cite{bottou2010large} as an optimization approach to train the network for 100K iterations, and the hyper-parameters of SGD, i.e., weight decay, momentum and gamma,  are set to be 0.0001, 0.9 and 0.1, respectively.

\begin{table*}
\scriptsize
\centering
\caption{Performance comparison in terms of AP with state-of-the-art methods on DOTA. FCOS-O means the results from our baseline method FCOS-O, RoI Trans means the two-stage method with RoI transfer in \cite{ding2018learning}. IENet is our proposed model, and `-T' determines the schedule with our proposed `geometric transformation' method, and `-A' are methods with attention process, `-D' means deformable convolution method proposed in \cite{dai2017deformable}. }
\begin{center}
\begin{tabular}{ccccccccccccccccc}
\hline
{}    & mAP & Plane & BD & Bridge & GTF & SV & LV & Ship & TC & BC & ST & SBF & RA & Harbor & SP & HC  \\
\hline
\textbf{Two Stage}	\\
\hline
RoI Trans    & \textbf{69.56}    & \textbf{88.64}      &\textbf{78.52}    &\textbf{43.44}        &\textbf{75.92}     &\textbf{68.81}   &\textbf{73.68}    &\textbf{83.59}      & \textbf{90.74}    & \textbf{77.27}    &\textbf{81.46}    & \textbf{ 58.39}    &\textbf{ 53.54}    &  \textbf{ 62.83}      &  58.93   &   47.67   \\

FR-O      & 54.13    & 79.42      & 77.13   & 17.70       & 64.05    & 35.30   & 38.02   & 37.16     & 89.41   & 69.64   & 59.28   & 50.30    & 52.91   & 47.89       &47.40    & 46.30    \\

R2PN \cite{zhang2018toward}    & 61.01   &80.94    & 65.75   & 35.34    & 67.44    & 59.92    & 50.91    & 55.81    & 90.67    & 66.92    & 72.39    & 55.06    & 52.23    & 55.14    & 53.35    & 48.22    \\

RC2 \cite{liu2017rotated}  & 60.67   & 88.52    & 71.2    & 31.66    & 59.3    & 51.85    & 56.19    & 57.25    & 90.81    & 72.84    & 67.38    & 56.69    & 52.84    & 53.08    & 51.94    & 53.58    \\

R-DFPN \cite{yang2018automatic}    &  57.94   & 80.92    & 65.82    & 33.77    & 58.94    & 55.77    & 50.94    & 54.78    & 90.33    & 66.34    & 68.66    & 48.73    & 51.76    & 55.1    & 51.32    & 35.88    \\
\hline
\textbf{One Stage}	\\
\hline
FCOS-O         & 46.58    & 84.41      &60.65    &22.37        &31.59     &51.17   &45.71    &60.46      &87.09    &42.28    &65.74    & 19.79    &23.04    &  34.22      & 39.10   &  34.15   \\

IENet(w/o -A, -D)       & 57.92     & 86.41   & 75.06       & 25.05    & 47.95   & 57.88   & 58.04     & 69.95   & \textbf{90.17}   &\textbf{75.98}   & 70.13    & 38.35   & 36.64       &46.13    & 56.85   & 34.19 \\

IENet(w/o -A)   & 58.91    & 84.87      & \textbf{75.97}   & 31.24       & 47.15    &57.74    & 57.64   & 69.67     & 90.16   & 63.75   & \textbf{76.31}   & \textbf{45.62}    & \textbf{41.89}   & 42.15       & 57.68   &  41.78   \\

IENet(w/o -D)     &  61.13    & 87.79    &  70.87     & \textbf{34.52}    &  51.38      & 63.88    & 65.45   & 71.11   &  89.85    & 70.50   & 73.86   & 37.49    & 41.40    & \textbf{50.44}   &  60.39      &48.07   \\

IENet        &  \textbf{61.24}    & \textbf{88.15}    &  71.38     & 34.26    &  \textbf{51.78}     & \textbf{63.78}   &\textbf{65.63}   & \textbf{71.61}   &  90.11    & 71.07   & 73.63   & 37.62    & 41.52    & 48.07   &  \textbf{60.53}      &\textbf{49.53}   \\
\hline
\end{tabular}
\end{center}
\label{table:1}
\end{table*}

\subsection{Evaluation and Comparisons with State-of-the-art Methods}
We compare the proposed IENet with state-of-the-art one and two-stage detectors, in terms of detection accuracy and efficiency. As illustrated in Fig.~\ref{model} (b), we modify FCOS by adding a convolutional layer in the regression branch to enable FCOS to regress the orientation parameters directly, which is used as our baseline FCOS-O. In addition, two top-performance two-stage orientation detectors RoI Transformer~\cite{ding2018learning} and Faster R-CNN OBB detector~\cite{xia2018dota} are also employed for comparison. The results on the DOTA and HRSC2016 datasets are shown in Table~\ref{table:1} and Table~\ref{table:2}, respectively.

To demonstrate the effectiveness of the modules (i.e., geometric transformation, IE module, DCN convolutional layer) used in our proposed framework, we also use different combinations of these modules as our baseline methods. The results are also given in Table~\ref{table:1}. 
Here, the IENet contains all the modules, specifically geometric transformation, IE module, and DCN. The components in the IENet framework  are indicated in abbreviated form, namely, `-A' indicates the IE module, and `-D' indicates DCN convolutional layer. Moreover, w and w/o  are used to indicate with and without specified modules respectively. In this way, the results can reveal the underlying effect of different modules combinations, and the detail is described in Section~\ref{ienet_ablation_studies}. Moreover, some qualitative results on both datasets are shown in Fig. \ref{DOTA_results} and Fig. \ref{hrsc_results},  which demonstrate the robust and accurate detection performance of our proposed IENet in both dense and sparse scenarios.

\begin{table}[ht]
\small
\centering
\caption{Performance comparison in terms of mAP with state-of-the-art methods on HRSC2016, where RoI Trans means the  method with RoI transfer in~\cite{ding2018learning}, FR-O denotes the orientation Faster R-CNN, and FCOS-O is the baseline method.}
\begin{tabular}{cccccc}
\hline
{} & RoI Trans & RC2 \cite{liu2017rotated}   & R2PN \cite{zhang2018toward}  & FCOS-O         & IENet   \\
\hline
mAP     & 86.20     & 75.7  & 79.6      &  68.56       &  75.01         \\
\hline
\end{tabular}
\label{table:2}
\end{table}

\textbf{Evaluations on DOTA} As can be seen from Table \ref{table:1}, our proposed IENet based models achieve mAP score of $57.92$, $58.91$, $61.13$ and $61.24$, respectively. The proposed IENet not only significantly outperforms our baseline one-stage model FCOS-O, but also exceeds the two-stage models. When compared with the one-stage detector, our proposed IENet achieves an mAP score of $61.24$, whereas the result from FCOS-O is $46.58$, and our IENet outperforms the FCOS-O by 14.66 points. When compared with two-stage detectors, our  IENet models can outperform most state-of-art two-stage detectors except RoI Trans. In addition, even our worst model (i.e., IENet w/o -A, -D) can achieve $57.92$ mAP score, which still outperforms FR-O by 3.79 points.
Nevertheless, our IENet model can achieve better performance on 2 categories (i.e., SP, HC) in the DOTA dataset with fewer network parameters and less computational complexity.

\textbf{HRSC2016 Evaluation} Table~\ref{table:2} shows the evaluation results on the HRSC2016 dataset.
Although the performance of the proposed IENet is worse than the two-stage methods,  it still achieves $75.01$ mAP score, which outperforms the one-stage baseline method FCOS-O by 6.45 points. Our IENet performs less well as compared with two-stage  methods on the HRSC2016 dataset, for the following two reasons: (1) Most ships in HRSC2016 are berthed nearshore, thus can hardly be separated from the terminal. However, unlike two-stage detectors, the one-stage detectors do not use RoI to extract foreground from background, thereby affecting the performance.
(2) Instead of using RoI, one stage detectors use predictions with high confidence as final results, which require a large number of samples to achieve optimal performance. However, HRSC2016 only contains few samples, hence one-stage detectors, such as IENet, potentially overfit with the training set.
Nevertheless, we think our model achieves competitive performance when compared with the two-stage model RC2.

\textbf{Speed-accuracy Comparison} In order to evaluate the efficiency of the proposed IENet, training time (tr-time), inference time (inf-time) and size of parameters (params) are evaluated on DOTA, and the result is given in Table~\ref{table:3}. Specifically, when compared with the anchor-free one-stage detector FCOS-O, IENet can achieve substantial improvement on accuracy while maintaining low complexity and light-weight. When compared with the two-stage detectors, although IENet is not always better in accuracy, it has advantages in efficiency and model size.
\begin{table}
\small
\centering
\caption{Speed-accuracy trade-off comparison for IENet, where RoI Trans denotes the method with RoI transfer in \cite{ding2018learning}, FR-O is the orientation Faster R-CNN model and the IENet is our IENet model. Tr-time denotes time for training, inf-time denotes inference time and Params denotes the total parameter numbers.}
\begin{tabular}{ccccc}
\hline
{}   & mAP & tr-time(s) & inf-time(s) & params(MB)            \\
\hline
RoI Tran & 69.56    & 0.236            & 0.084           & 273                 \\

FR-O      & 54.13    & 0.221            & 0.102           & 270                  \\

FCOS-O   & 46.58    & 0.109            & 0.056           & 208                 \\

IENet     & 61.24    & 0.111            & 0.059           & 212                  \\
\hline
\end{tabular}
\label{table:3}
\end{table}

\begin{figure*}[t]
	\begin{center}
		\includegraphics[width=0.8\textwidth]{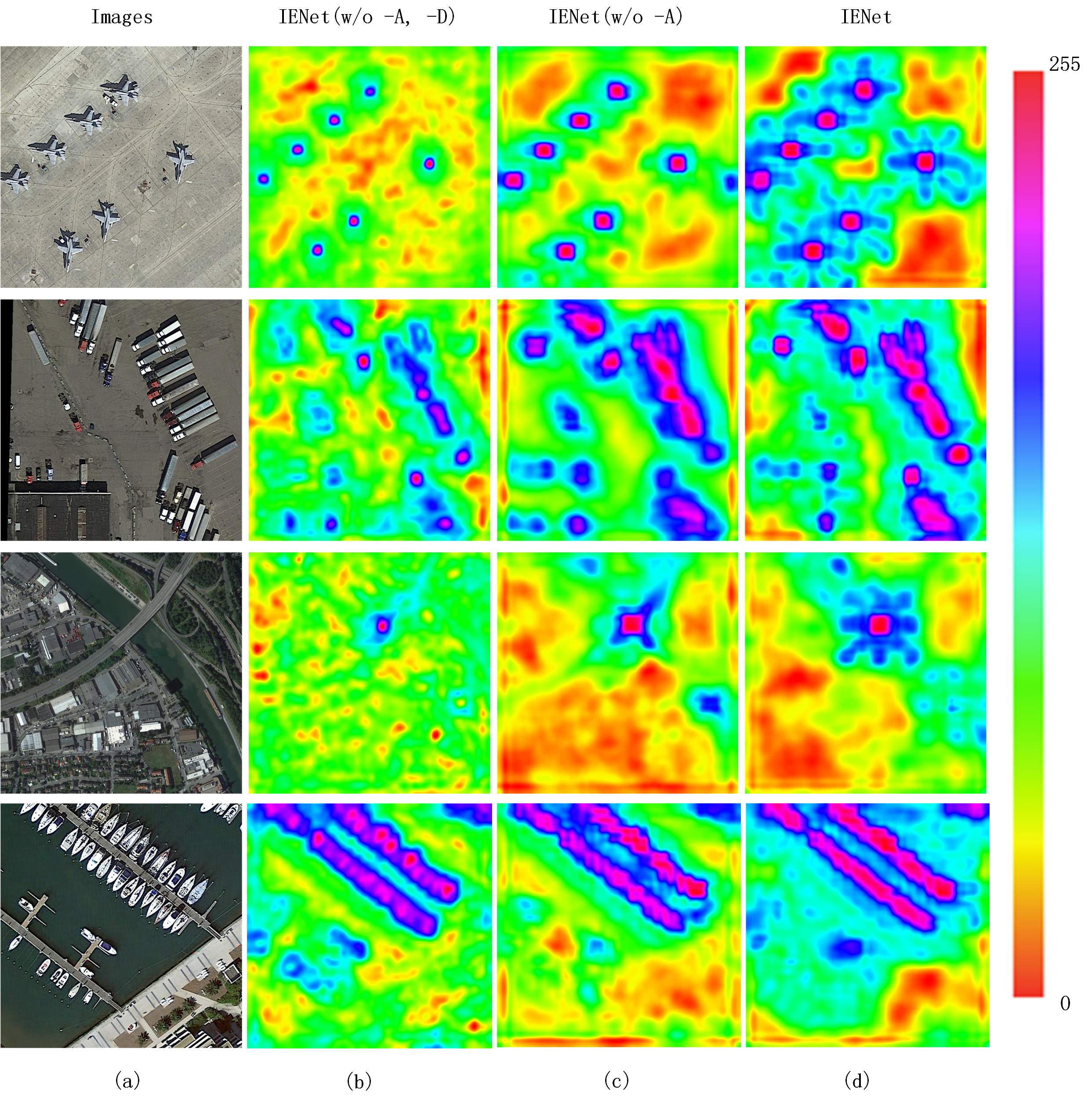}
	\end{center}
	\caption{Feature visualization comparison for IENet for different targets. Here IENet is our proposed architecture, where self-attention mechanism is employed to fuse feature maps from classification and box regression branches onto orientation regression branches. The colour map is drawn using HSV colour space and the gradient-weighted class activation map is using to visualize the feature maps.}\label{figure:7}
\end{figure*}

\subsection{IENet Ablation Studies}\label{ienet_ablation_studies}
To illustrate the contributions of the main components in our method, i.e., geometric transformation and the IE module, we perform ablation studies on these components using the DOTA dataset. The results are given in Table~\ref{table:4}, where we employ mAP as the performance metric.

\begin{table}[h]
\centering
\caption{Ablative experiments for the IE module on the DOTA dataset. We use ResNet-101 for all experiments. The results show the improvement of our proposed geometric transformation and IE module.}
\begin{tabular}{ccccc}
	\hline
	{}     & DCN      & Geo trans     & IE Module       & mAP            \\
	\hline
	FCOS-O   & {}    & {}         & {}           & 46.58                 \\
	
	FCOS-O(w -D)   & $\checkmark$    & {}         & {}           & 47.39                 \\
	\hline
	IENet(w/o -A, -D)   & {}    & $\checkmark$        & {}            & 57.92                  \\

	IENet(w/o -A)   & {}    & $\checkmark$        & {}            & 58.91                  \\

	IENet(w/o -D)   & {}    & $\checkmark$         & $\checkmark$           & 61.13                  \\

	IENet    & $\checkmark$    & $\checkmark$         & $\checkmark$           & 61.24  \\
	\hline
\end{tabular}
\label{table:4}
\end{table}

From Table \ref{table:4}, we can see that, by using the proposed geometric transformation process, IENet(w/o -A, -D) can significantly improve the accuracy of the baseline model FCOS-O, and adding the self-attention based IE module, i.e., IENet(w/o -D) can further improve the accuracy, which outperforms IENet(w/o -A, -D). However, by employing the DCN, the performance of our IENet only has slight improvement when compared to the model without DCN process, i.e., IENet(w/o -D). Here, the details of the ablation studies are described as follows.

\textbf{Geometric Transformation} The geometric transformation is proposed to address the OBB regression problem by using an HBB and its corresponding orientation parameters, where six parameters are used for prediction. As shown in Table~\ref{table:4}, by employing the proposed geometric transformation, IENet(w/o -A, -D) achieves 57.92 mAP score, which is $11.34$ points higher than the baseline FCOS-O (i.e., $46.58$). Thus, the result shows that the geometric transformation facilitates the prediction of the orientations of the objects.

\textbf{Interactive Embranchment Module} In Table~\ref{table:4}, we find that FCOS-O(w -D) can achieve a mAP score of $47.39$, due to the help of DCN, which is a slight improvement over the use of the FCOS-O model.
In addition, we observe that the DCN helps IENet(w/o -A) achieve a mAP score of 58.91, which is 1 point higher than IENet(w/o -A, -D). These results show that DCN can generally improve model accuracy. Moreover, the model IENet(w/o -D) achieves an mAP score of $61.13$, due to the use of the IE module, which is $3.21$ and $2.22$ points higher than IENet(w/o -A, -D) and IENet(w/o -A), respectively. The results show that the model with IE module can not only improve IENet(w/o -A, -D), but also outperform IENet(w/o -A). However, when employing DCN in IENet(w/o -D),  the mAP score achieved is 61.24, which provides only 0.11 point improvement as compared with IENet(w/o -D). These results indicate that our IE module already extracts the befitting features for the final prediction. It means, with or without DCN, our models IENet(w/o -D) and IENet can obtain nearly the same performance, the performance improvement mainly comes from the use of the IE module.
\subsection{Feature Map Visualization}
\label{sec:4_5}
To further understand the effect of the IE module on our model, we visualize the classification branch feature maps with respect to the predicted class. The feature visualization comparison is provided  among our three models, which are
IENet without IE module and DCN, IENet without IE module and IENet. The result is illustrated in Fig.~\ref{figure:7}.

\textbf{Geometric Transformation} From Fig.~\ref{figure:7} (b), the colour map shows that each object in the images has a small area of high gradient value, which means the network can find the object in the images, but lacks confidence in predicting the object. Moreover, the colour map presents a large area of green part outside the object, which indicates that the network is not fully confident in excluding the background.

\textbf{Deformable Convolution Network} From Fig.~\ref{figure:7} (c), the DCN method used here is served as an attention module, which can alleviate the low confidence problem that we mentioned above. As a result, the DCN increases the network attention on objects and reduces the network attention on background. However, the DCN only boosts the classification performance by giving more attention on the center of the object, and cannot provide the orientation information about the object.

\textbf{Interactive Embranchment Module} From Fig.~\ref{figure:7} (d), our best model IENet can outperform  IENet(w/o -A) by using the IE module. The IENet not only can resolve the two major problems  as  shown in the visualize colour maps Fig.~\ref{figure:7}  (b) and Fig.~\ref{figure:7}  (c), but also can provide more attention around the object, which forces the network to look around the object and understand the orientation of the object. Therefore, the use of the IE module can guide the network to pay more attention on target orientation. Hence, it not only boosts the classification performance, but also provides the orientation information. This is because  the interactive behavior happened in the IE module among three prediction branches. However, this behavior may also bring interference and noise problem to the network, which we will mention next in sub-Section \ref{sec:4_6}.

\subsection{Potential Study}
\label{sec:4_6}
In this work, we present a new method (i.e., geometric transformation) to express the orientation bounding boxes by using their surrounding horizontal bounding boxes and  orientation parameters. In this way, the ground truth HBB used for training is achieved by the labeled OBB with an inverse geometric transformation, which contains more background than HBB given by ground truth information. Moreover, the background will be regarded as part of the target during training, and reduce the prediction accuracy.

Despite the lower accuracy in the OBB prediction, we still maintain a well predicted outer box for the sparsely distributed targets by using the combination of the smooth-l1 and IoU loss. Nevertheless, the prediction accuracy for crowded targets needs to be improved. Therefore, we think that is the reason why our model is unable to achieve high accuracy for some crowded targets, such as  small vehicles in Table \ref{table:1}.

\section{Conclusion}
\label{sec:conclusion}
In this paper, a one-stage orientation detector, IENet, was proposed for orientation target detection in aerial images with an anchor free solution. A one-stage anchor-free keypoint based architecture was employed as the baseline framework, where the orientation prediction problem was addressed by employing a novel geometric transformation method. Moreover, a self-attention mechanism based IE model was used to combine features for orientation parameters prediction. Comparison results showed  improvement in terms of accuracy as well as the efficiency of our proposed IENet. Ablation studies resolved the excellent performance from each component of IENet. In our future study, a more robust feature interacting method will be sought to extract better features for OBB prediction, then a more stable rotation region of interest method will be employed to mitigate the background area and yield state-of-the-art performance.

\section*{Acknowledgments}
This work was partly supported by the Fundamental Research Funds for the Central Universities under Grant No. 3072020CFT0602, the Open Research Fund of State Key Laboratory of Space-Ground Integrated Information Tech- nology under Grant No. 2018\_SGIIT\_ KFJJ\_ AI\_01, and the National Natural Science Foundation of China under Grant No. 61806018.
\IEEEpeerreviewmaketitle

\bibliographystyle{IEEEtran}
\bibliography{refs_IENet}

%

%
%
%




\end{document}